\pdfoutput=1

\documentclass[letterpaper]{article}

\usepackage{rotating}
\usepackage{pdflscape}

\usepackage{times}
\usepackage{helvet}
\usepackage{courier}
\newenvironment{definition}

\usepackage[numbers,sort&compress,longnamesfirst,sectionbib]{natbib}

\usepackage[labelfont=bf,font=footnotesize]{caption}  % for hanging captions

\usepackage{amsmath}

\usepackage[percent]{overpic}

\usepackage{xspace}
\usepackage{balance}

\usepackage{enumitem} % for setlist space

\newcommand{\MVC}{\textsc{MVC}\xspace}

\newcommand{\FastVC}{\textsc{FastVC}\xspace}
\newcommand{\NuMVC}{\textsc{NuMVC}\xspace}
\newcommand{\CLK}{\textsc{CLK}\xspace}
\newcommand{\restarts}[2]{\textsc{Restarts}$^{\text{#1}}_{\text{#2\%}}$}
\newcommand{\restartsluby}[2]{\textsc{RestartsLuby}$^{\text{#1}}_{\text{#2\%}}$}

\usepackage{xcolor}

\usepackage[bookmarks=false]{hyperref}

\newcommand{\ignore}[1]{}

\usepackage{tikz}
\usepackage{pgfplots}
\usepackage{pgfplotstable}
\usepackage{colortbl}
\usepackage{tabularx}
\usetikzlibrary{shapes,decorations.shapes}
\pgfplotsset{compat=newest}

\pgfplotstableset{
display columns/0/.style={column type={>{\centering\arraybackslash}p{8mm}}},%column width from http://tex.stackexchange.com/questions/236603/how-can-i-wrap-the-text-in-certain-columns-using-pgfplotstable
    /color cells/min/.initial=0,
    /color cells/max/.initial=1000,
    /color cells/textcolor/.initial=,
    color cells/.code={%
        \pgfqkeys{/color cells}{#1}%
        \pgfkeysalso{%
            postproc cell content/.code={%
                \begingroup
                \pgfkeysgetvalue{/pgfplots/table/@preprocessed cell content}\value
                \ifx\value\empty
                    \endgroup
                \else
                \pgfmathfloatparsenumber{\value}%
                \pgfmathfloattofixed{\pgfmathresult}%
                \let\value=\pgfmathresult
                \pgfplotscolormapaccess
                    [\pgfkeysvalueof{/color cells/min}:\pgfkeysvalueof{/color cells/max}]
                    {\value}
                    {\pgfkeysvalueof{/pgfplots/colormap name}}%
                \pgfkeysgetvalue{/pgfplots/table/@cell content}\typesetvalue
                \pgfkeysgetvalue{/color cells/textcolor}\textcolorvalue
                \toks0=\expandafter{\typesetvalue}%
                \xdef\temp{%
                    \noexpand\pgfkeysalso{%
                        @cell content={%
                            \noexpand\cellcolor[rgb]{\pgfmathresult}%
                            \noexpand\definecolor{mapped color}{rgb}{\pgfmathresult}%
                            \ifx\textcolorvalue\empty
                            \else
                                \noexpand\color{\textcolorvalue}%
                            \fi
                            \the\toks0 %
                        }%
                    }%
                }%
                \endgroup
                \temp
                \fi
            }%
        }%
    }
}

\title{A Generic Bet-and-run Strategy for Speeding Up Traveling Salesperson and Minimum Vertex Cover\footnote{This work has been supported by the ARC Discovery Early Career Researcher Award DE160100850.}
}

\author{%(author details removed for double-blind reviewing)
Tobias Friedrich$^1$, Timo K\"{o}tzing$^1$, Markus Wagner$^2$\\
$^1$Hasso Plattner Institute, Potsdam, Germany\\
$^2$Optimisation and Logistics, The University of Adelaide, Adelaide, Australia\\
\ \\
\textit{Note: this article is currently under review.}
}

\begin{document}

\maketitle

\sloppy

\begin{abstract}
A common strategy for improving optimization algorithms is to restart the algorithm when it is believed to be trapped in an inferior part of the search space. However, while specific restart strategies have been developed for specific problems (and specific algorithms), restarts are typically not regarded as a general tool to speed up an optimization algorithm. In fact, many optimization algorithms do not employ restarts at all.

Recently, \emph{bet-and-run} was introduced in the context of mixed-integer programming, where first a number of short runs with randomized initial conditions is made, and then the most promising run of these is continued. In this article, we consider two classical NP-complete combinatorial optimization problems, traveling salesperson and minimum vertex cover, and study the effectiveness of different bet-and-run strategies. In particular, our restart strategies do not take any problem knowledge into account, nor are tailored to the optimization algorithm. Therefore, they can be used off-the-shelf. We observe that state-of-the-art solvers for these problems can benefit significantly from restarts on standard benchmark instances.
\end{abstract}

\section{Introduction}

When a desktop PC is not working properly, the default answer of an experienced
system administrator is restarting it. The same holds for stochastic algorithms
and randomized search heuristics: If we are not satisfied with the result, we might
just try restarting the algorithm again and again. While thish is well-known~\citep{Mar:bc:03,Lou-Mar-Stu:bc:10}, very few algorithms directly incorporate such restart strategies. We assume that this is due to the added complexity of designing an appropriate restart strategy that is advantageous for the considered algorithm.

Hence, it would be beneficial to have a generic framework for restart strategies which is not overly dependent on the exact algorithm used or the problem under consideration. In this paper we want to show that there are restart strategies which are of benefit in a variety of settings.

There are some theories on how to choose optimal restart strategies, independently of the setting. For example, \citet{luby1993} showed that, for Las Vegas algorithms with known run time distribution, there is an optimal stopping time in order to minimize the expected running time. They also showed that, if the distribution is unknown, there is an universal sequence of running times given by (1,1,2,1,1,2,4,1,1,2,1,1,2,4,8,...), which is the optimal restarting strategy up to constant factors. These results have the appeal that they can be used for every problem setting; however, they only apply to Las Vegas algorithms.

For the case of optimization, the situation is much less clear, with plenty of different approaches presented by the stochastic optimization community. A gentle introduction to practical approaches for such restart strategies is given by~\citet{Mar:bc:03} and \citet{Lou-Mar-Stu:bc:10}, and a recent theoretical result is presented by~\citet{Sch-Tey-Tey:c:12}. Particularly for the satisfiability problem (SAT), there are several studies that make an empirical comparison of a number of restart policies~\citep{Biere08, Huang2007}. These show the substantial impact of the restart policy on the efficiency of SAT solvers. In the context of satisfiability problems this might be unsurprising as state-of-the-art SAT and CSP solvers often speed up their search by learning ``no-goods'' during backtracking~\cite{CireKS14}.

While classical optimization algorithms are often deterministic and thus cannot be improved by restarts (neither their run time nor their outcome will alter), many modern optimization algorithms, while also working mostly deterministically, have some randomized component, for example by choosing a random starting point. Thus, %in many practical applications, 
the initial solution often strongly influences the quality of the outcome. It follows that it is natural to do several runs of the algorithm. Two very typical uses for an algorithm with time budget $t$ are to (a) use all of time $t$ for a single run of the algorithm (single-run strategy), or (b) to make a number of $k$ runs of the algorithm, each with running time $t/k$ (multi-run strategy).

\begin{figure}[tb]
\centering\includegraphics[width=0.95\columnwidth]{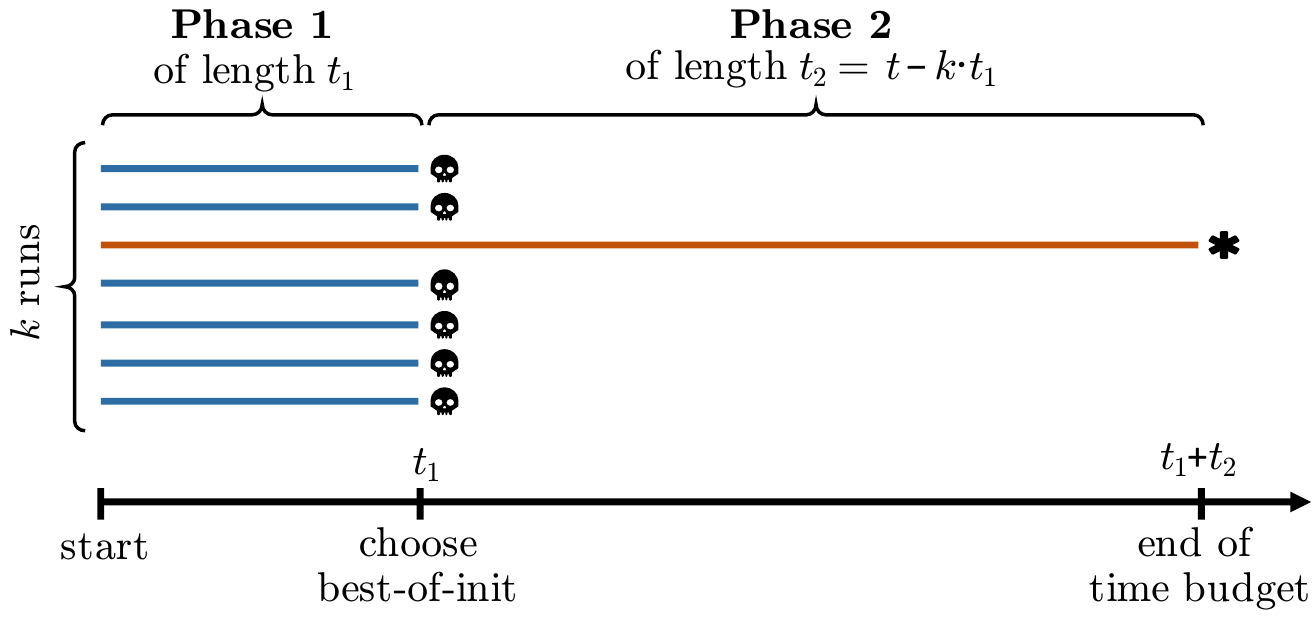}
\caption{Our \emph{bet-and-run} restart strategy starts with $k$ independent runs
and total time budget $t$.
After time $t_1$ all but the best run are terminated (marked with
\protect\includegraphics[height=2mm]{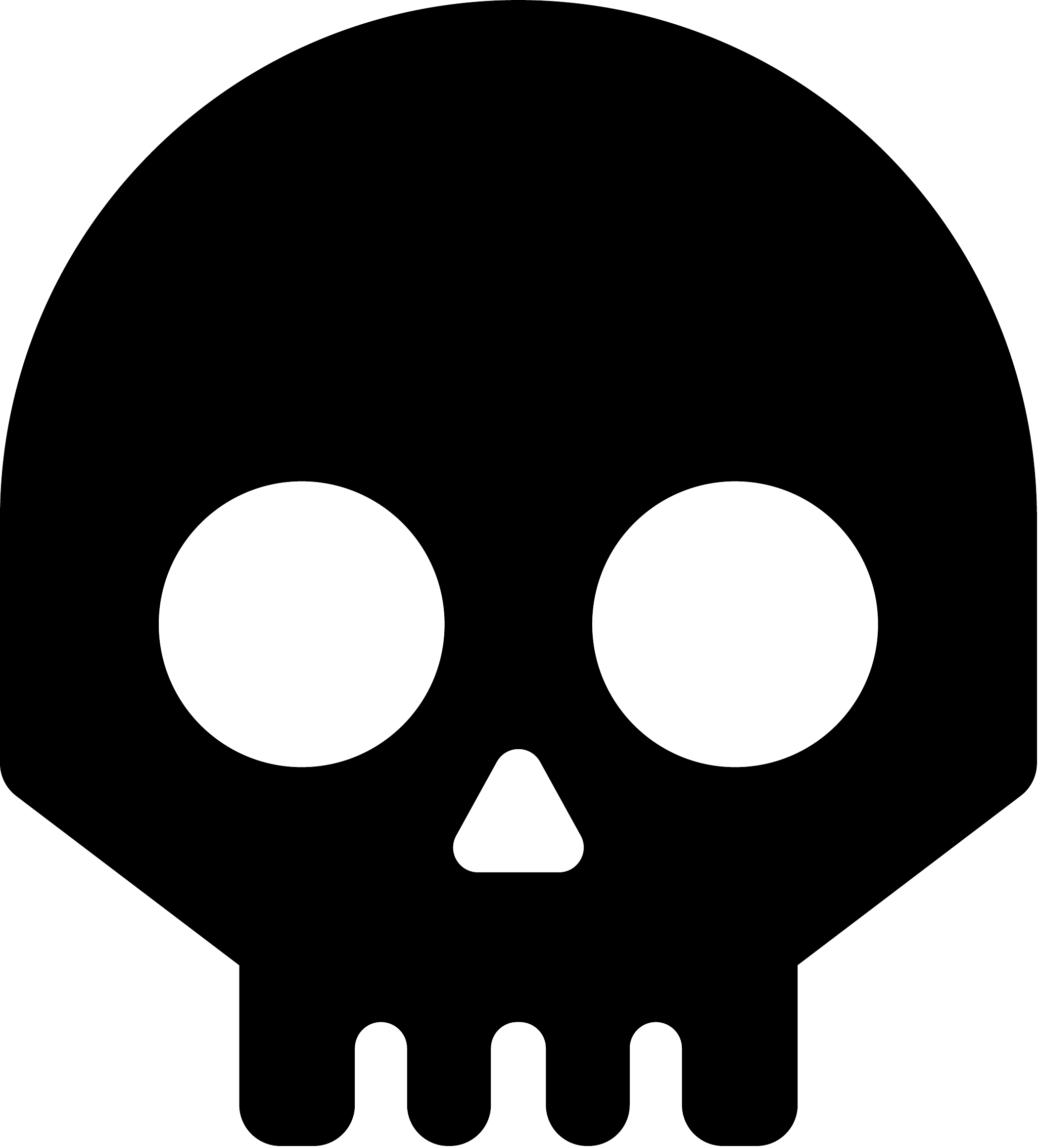}).
The best run (marked with
\protect\includegraphics[height=2.3mm]{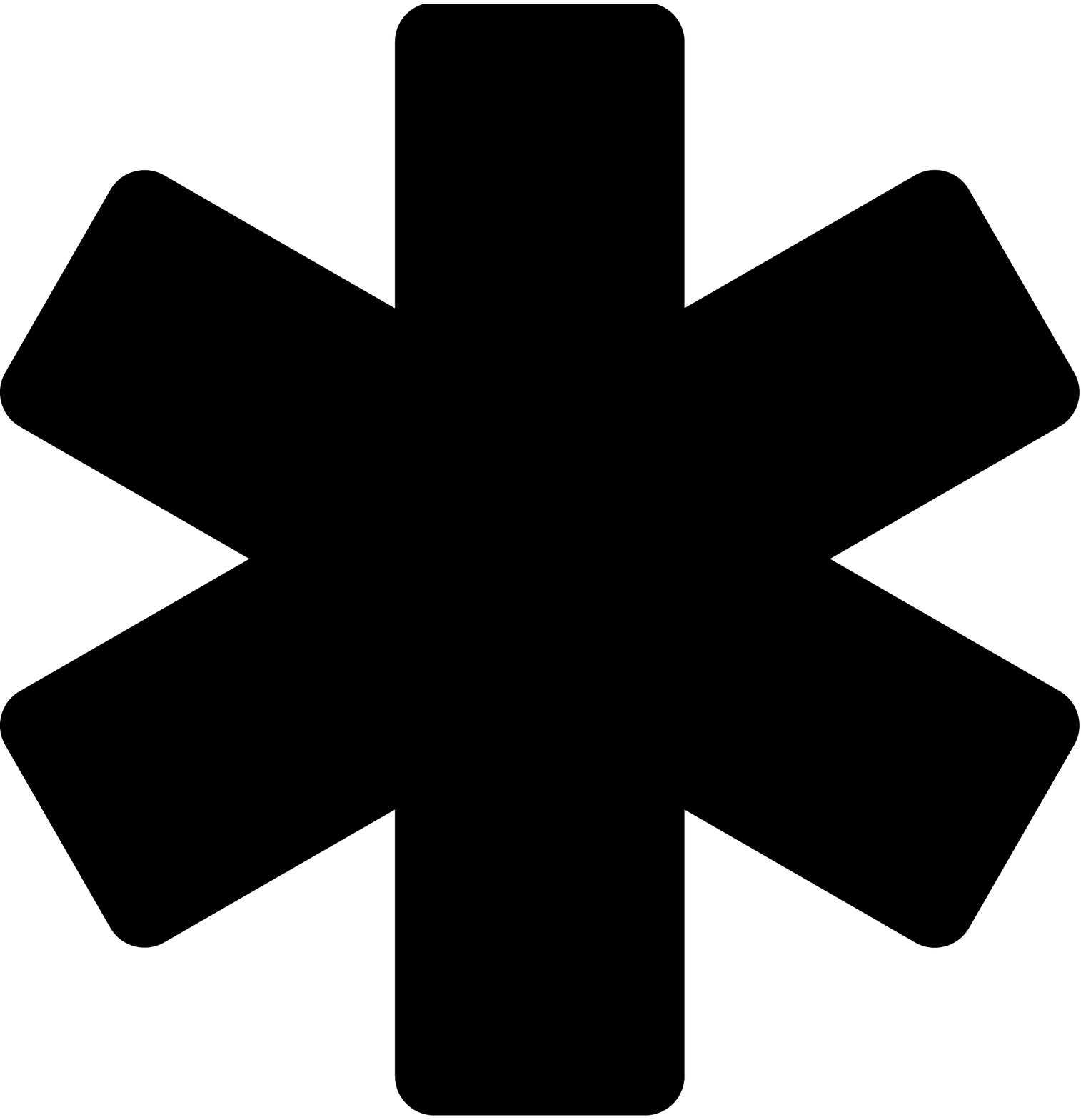})
continues for $t_2$ time steps until the total time budget runs out.
%Illustration of our bet-and-run restart strategy with $k$ runs; after time $t_1$ all but the best run are terminated.
}\label{fig:illustration}
\end{figure}

Extending these two classical strategies, 
\citet{Fischetti2014betandrun} investigated the use of the following \emph{bet-and-run} strategy with a total time limit $t$:
\begin{description}
\item[Phase 1] performs $k$ runs of the algorithm for some (short) time limit $t_1$ with $t_1\leq t/k$.
\item[Phase 2] uses remaining time $t_2=t-k\cdot t_1$ to continue \emph{only the best run} from the first phase until timeout.
\end{description}
This strategy is illustrated in Figure~\ref{fig:illustration}. Note that the multi-run strategy of restarting from scratch $k$~times is a special case by choosing $t_1=t/k$ and $t_2=0$ and the single-run strategy corresponds to $k=1$; thus, it suffices to consider different parameter settings of the \emph{bet-and-run} strategy to also cover these two strategies.

\citet{Fischetti2014betandrun} experimentally studied such a bet-and-run strategy for mixed-integer programming. They explicitly introduce diversity in the starting conditions of the used MIP solver (IBM ILOG CPLEX) by directly accessing internal mechanisms. In their experiments with $k=5$, bet-and-run was typically beneficial.
%
%a few very short runs are done and the most promising run was then brought to completion, with the result that bet-and-run is typically beneficial.
%
\citet{dePerthuisdeLaillevault2015onemaxInits} have recently shown that 
a \emph{bet-and-run} strategy can also benefit asymptotically from larger $k$: For the pseudo-boolean test function \textsc{OneMax} it was proven that choosing $k>1$ decreases the $O(n \log n)$ expected run time of the (1+1) evolutionary algorithm by an additive term of $\Omega(\sqrt{n})$~\citep{dePerthuisdeLaillevault2015onemaxInits}. They also rigorously showed that the optimal gain is achieved for some $k$ of order $k=\sqrt{n}$.

In this paper we want to show that there is \emph{no need} to tailor the restart strategy or to access the internal mechanisms of available solvers: in fact there are generic \emph{bet-and-run} restart strategies that consistently outperform single-run and multi-run strategies \emph{across different domains}! We benchmark our strategies on two different problems: traveling sales person (TSP) and minimum vertex cover (MVC). We observe statistically significant improvements of our bet-and-run strategy on standard corpora for state-of-the-art solvers for both optimization problems.

Details for our design choices can be found in Section~\ref{sec:ProblemsAndBenchmarks}, 
along with a formal definition of the problems.
Since it is a priori not obvious what \emph{bet-and-run} strategies are most promising, we define a generic scheme of restart strategies in Section~\ref{sec:RestartStrategies}; we compare these strategies in Section~\ref{sec:ChoosingStrategies}, where we find 14~parameter settings for the \emph{bet-and-run} strategy ($k$ and $t_1$) that are representative of the space of possible parameter settings. Finally, in Section~\ref{sec:final}, we show that there are \emph{bet-and-run} strategies that perform well across a wide range of instances in different domains.

\section{Problems and Benchmarks}
\label{sec:ProblemsAndBenchmarks}

In the following we briefly introduce the two NP-complete problems we consider, as well as the corresponding solvers and benchmarks used in this paper.

\subsection{Traveling Salesperson}

%Problem Def
The Traveling Salesperson problem considers an edge-weighted graph $G = (V,E,w)$, the vertices $V = \{1,\ldots,n\}$ are referred to as \emph{cities}. It asks for a permutation $\pi$ of $V$ such that
$$
\left(\sum_{i=1}^{n-1} w(\pi(i),\pi(i+1))\right) + w(\pi(n),\pi(1))
$$
(the cost of visiting the cities in the order of the permutation and then returning to the origin $\pi(1)$) is minimized.

%Motivation
Applications of the traveling salesperson problem arise naturally in areas like planning and logistics~\citep{Polacek2007823}, but they are also encountered in a large number of other domains; the textbook by \citet{applegate2011traveling} gives an overview of such encounters, listing areas as diverse as genome sequencing, drilling problems, aiming telescopes, and data clustering. TSP is identified as one of the most important (and most studied) optimization problems.

%Algo Def
We use the Chained-Lin-Kernighan (\CLK) heuristic~\citep{applegate2003chainedLK}, a state-of-the-art incomplete solver for the Traveling Salesperson problem.
The \CLK code is available online~\citep{urlCLK}. 
Despite being a few years old, \CLK still holds the records for a number of large TSPlib instances.

%Instances
The TSPlib is a classic repository of TSP instances~\citep{reinelt1991tsplib}, which are available online~\citep{urlTSPlib}.
For our first investigations, we pick from TSPlib the nine largest symmetric instances which have between 5,934 and 85,900 cities, and the Mona Lisa TSP Challenge instance~\citep{urlTSPMona}, which contains 100,000 cities. 
In summary, the instances are rl5934, pla7397, rl11849, usa13509, brd14051, d15112, d18512, pla33810, pla85900, and mona-lisa100k. 
For the first seven instances, \CLK takes less than 0.3 seconds to initialize. For the remaining three instances, the initialization times are 0.5, 2.5, and 2.5 seconds.

\subsection{Minimum Vertex Cover}

%Problem Def
Finding a minimum vertex cover of a graph is a classical NP-hard problem.
Given an unweighted, undirected graph $G = (V,E)$,
a vertex cover is defined as a subset of the vertices $S \subseteq V$,
such that every edge of $G$ has an endpoint in $S$, i.e. for all edges $\{u,v\} \in E$,
\[
u \in S\text{ or }v \in S.
\]
The NP-complete decision problem $k$-vertex cover decides whether
a vertex cover of size $k$ exists. We consider the optimization problem
which aims at finding a vertex cover of minimum size.

%Motivation
Applications of the vertex cover problem arise in various areas like network security, scheduling and VLSI design~\citep{Gomes2006}. To give an example,
finding a minimum vertex cover in a network corresponds to locating an optimal set of nodes on which to strategically place controllers such that they can monitor the data going through every link in the network.
The vertex cover problem is also closely related to the question of finding a maximum clique. This has a range of applications in bioinformatics and biology, such as identifying related protein sequences~\citep{Abu-Khzam2006}.

%Algo Def
%\Tobias{Hier fehlt besseres Argument, warum \FastVC}
Over the past two decades, numerous algorithms have been proposed for solving the vertex cover problem. We choose \FastVC~\citep{cai2015fastvc} over the popular \NuMVC~\citep{cai2013numvc} as a solver for the minimum vertex cover problem as it works better for massive graphs.
\FastVC is based on two low-complexity heuristics, one for initial construction of a vertex cover, and one to choose the vertex to be removed in each exchanging step.
The code of \FastVC is available online~\citep{urlFastVC}. 

%Instances
For our initial experimental investigations, we select from the 86 instances used by~\citet{cai2015fastvc} the 10 instances for which \FastVC outperforms \NuMVC the most. With this approach to instance selection (which differs from the one we used for TSP) we attempt to further increase the performance gap between both algorithms. 
In the order of increasing performance difference, these instances are rec-amazon, large-soc-gowalla, soc-digg, sc-shipsec1, soc-youtube, sc-shipsec5, soc-flickr, soc-youtube-snap, 
web-it-2004, and ca-coauthors-dblp. On all instances, \FastVC's initialisation takes at most $1.5$ seconds. All instances are available online~\citep{urlVC}.

\section{Restart Strategies}
\label{sec:RestartStrategies}

A restart strategy describes how the total time budget is distributed over a number of independent runs. We consider two different kinds of restart strategies as follows. On the one hand, we consider the \emph{bet-and-run} strategies where all initial runs have the same length. On the other hand, we are inspired by~\citet{luby1993} for defining a kind of restart strategy with different lengths in an attempt to be more robust with respect to choosing the right time $s$ for the initial runs. \citet{luby1993} define a simple universal strategy, defined as an (infinite) sequence indicating how many time units should be used for each run. This sequence is given as
$$
S^\text{univ}=(1,1,2,1,1,2,4,1,1,2,1,1,2,4,8,1,\ldots)
$$
and, more formally, by $\forall i \geq 1$,
$$
S^\text{univ}(i) = 
\begin{cases} 
  2^{k-1}, 											& \text{if } i=2^k - 1;\\
  S^\text{univ}(i-2^{k-1}+1), 	& \text{if } 2^{k-1} \leq i < 2^k-1.
\end{cases}
$$
The numbers given in the Luby sequence refer to the number of time units to employ per run. Thus, in order to use the Luby sequence, we have to define the length of this time unit, which we will call \emph{Luby time unit}. We indicate it as $x$\% of the total time budget. Thus, we get the following definition.

\begin{definition}
\label{def:restarts}

\restarts{k}{x} refers to the strategy where $k$ initial runs are performed, and each of the runs has a computational budget of $x$\% of the total time budget.

\restartsluby{k}{x} refers to the strategy that uses in its first phase runs whose lengths are defined by the Luby sequence. $k$ refers to the sequence length used in the first phase, and each Luby time unit is $x$\% of the total time.
\end{definition}

\section{Choosing Representative Restart Strategies}
\label{sec:ChoosingStrategies}

\newlength{\heatmapheight}
\setlength{\heatmapheight}{60mm}

\begin{landscape}
\begin{figure*}[t]%
\centering%
\includegraphics[height=\heatmapheight,trim={0 0 11.5mm 0},clip]{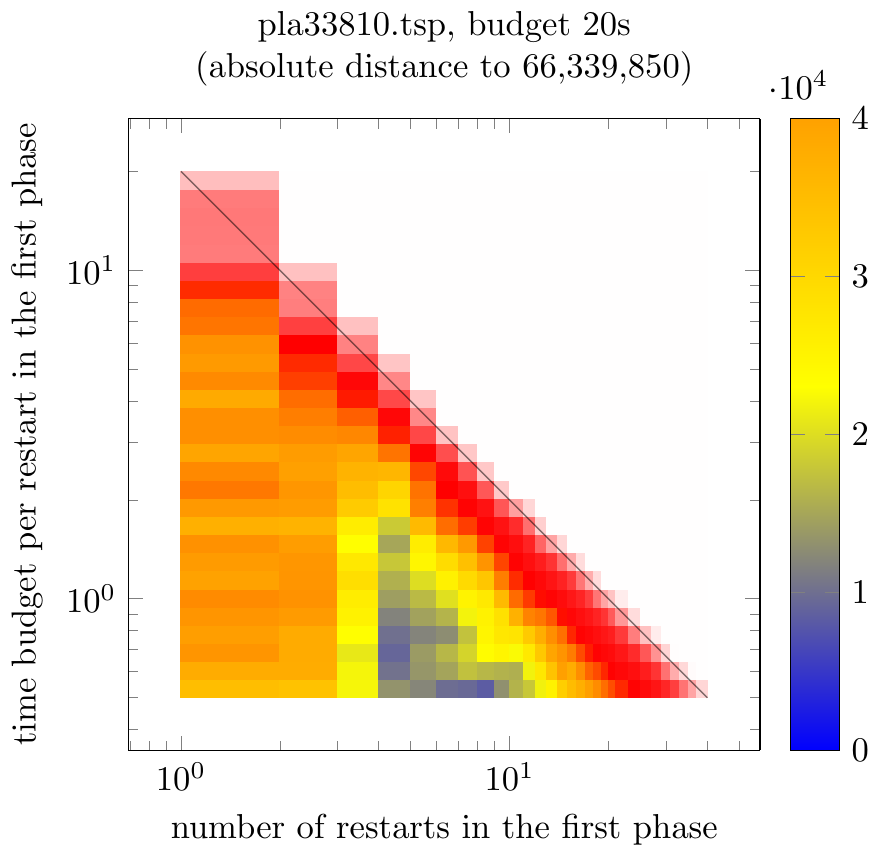}
\includegraphics[height=\heatmapheight,trim={4.5mm 0 11.5mm 0},clip]{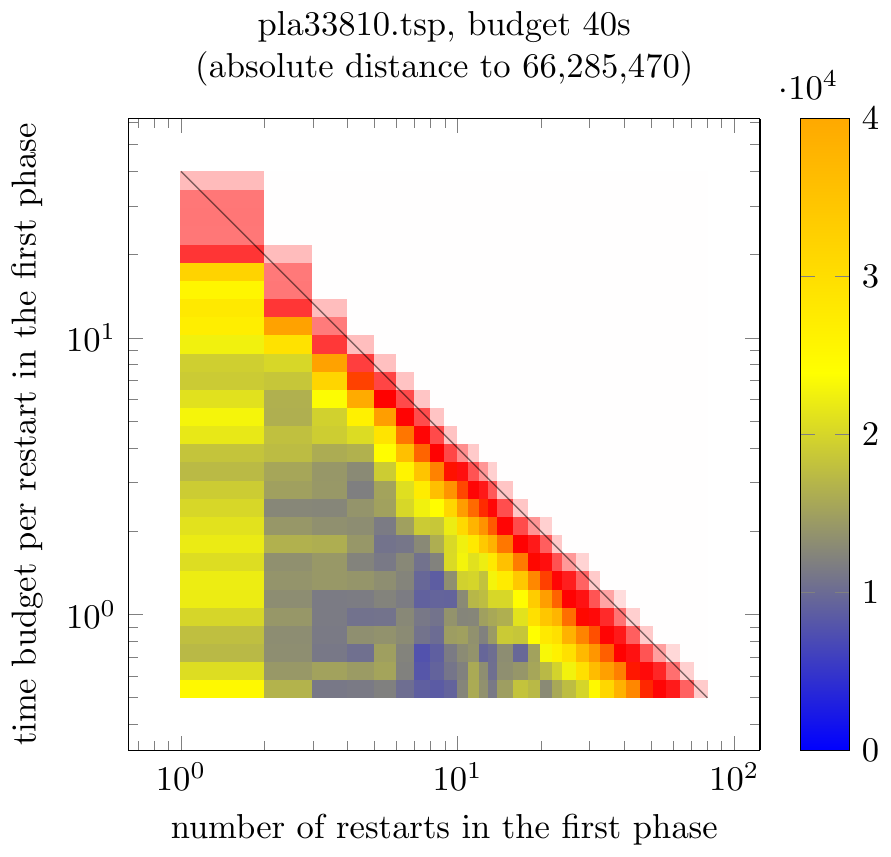}
\includegraphics[height=\heatmapheight,trim={4.5mm 0 0 0},clip]{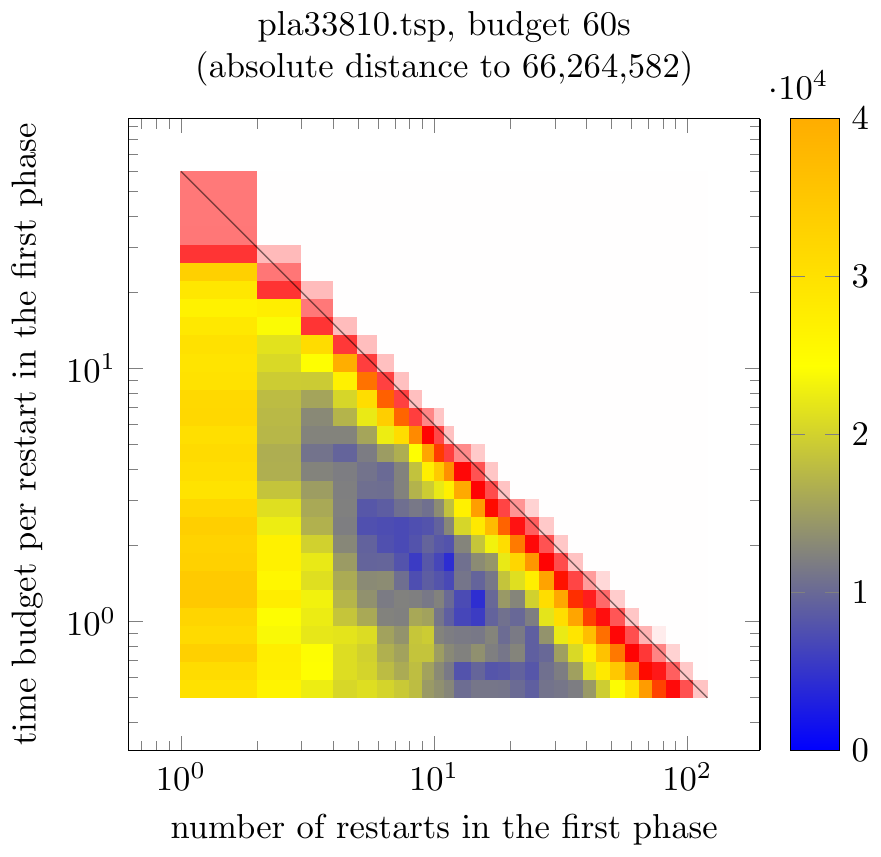}
\caption{Comparison of different restart strategies \restarts{$k$}{$x$}
and total time budgets for the TSP instance pla33810.
The plots show the average quality of the solution discovered with restart strategy \restarts{$k$}{$x$}
compared to the best average found in all runs (smaller values are better). The best average is shown in the title of each heatmap.
For example in the leftmost plot, the $x$-axis shows the number of runs~$k=1\ldots40$ in the first phase, and the $y$-axis shows the time budget per restart in the first phase in seconds. %~$x=0.5\%\ldots 20\%$.
The black diagonal is the line $k\cdot x=100\%$ of regular full-restart strategies
with no best-of phase.
The color is chosen depending on the average distance to the best average of 100~independent repetitions,
and the cells are colored based on the average of the corners.
In summery, we observe that restart strategies which perform a few short runs % (e.g. \restarts{$10$}{$5$})
perform
better on average than e.g.~no restarts (=only one run) or full-restarts (=diagonal line).
}
\label{fig:heatmapTime}
\end{figure*}
\end{landscape}

\setlength{\heatmapheight}{51mm}

\begin{figure*}[t]
\centering
\includegraphics[height=\heatmapheight,trim={0 0 0 0mm},clip]{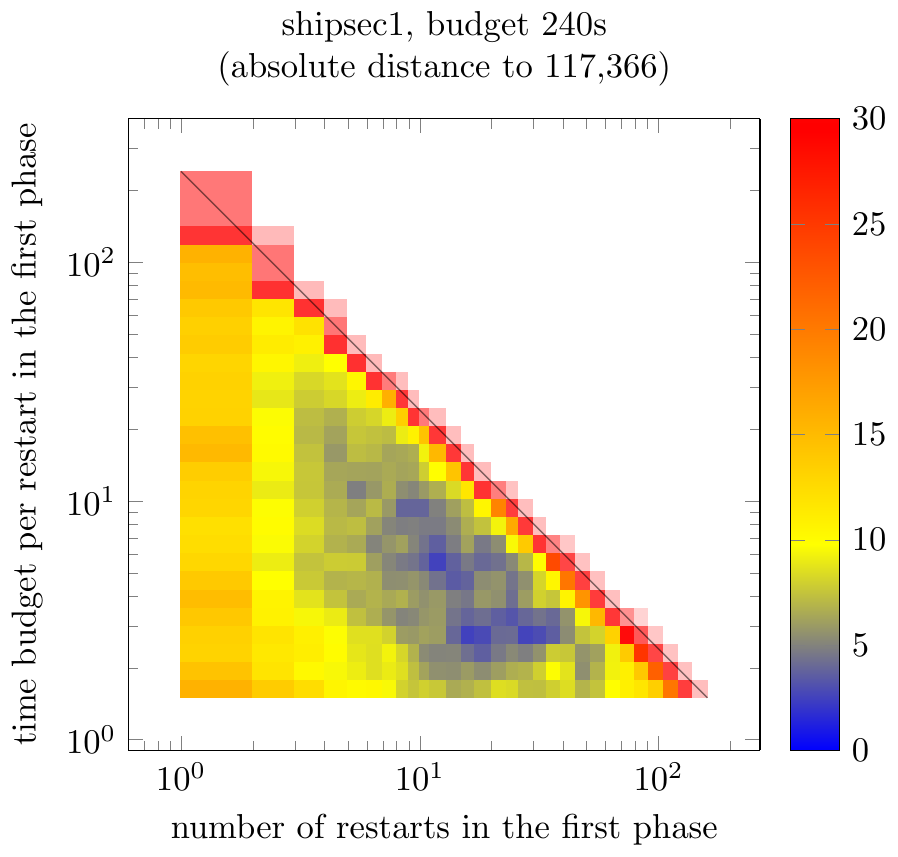}\hspace{5mm}
\includegraphics[height=\heatmapheight,trim={0 0 0 0mm},clip]{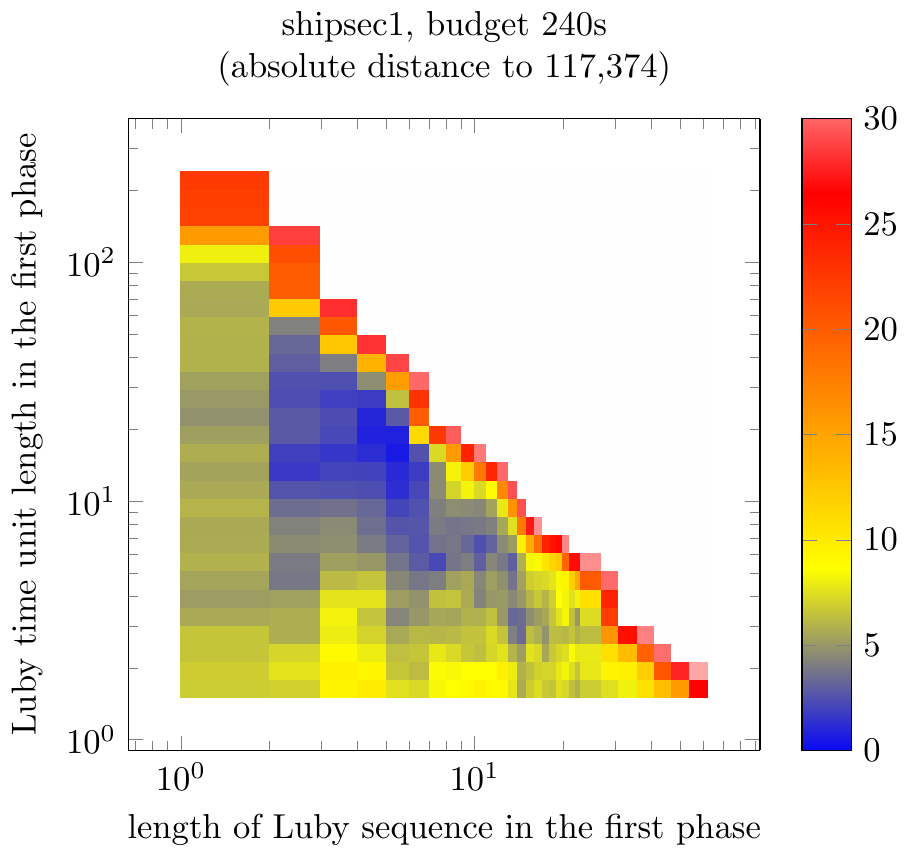}
\caption{For %the \MaxSAT instance scpcyc11 (top row) and 
the MVC instance shipsec1, %(bottom row)
we show the \textsc{Restarts} schemes on the left, and the \textsc{RestartsLuby} scheme on the right. 
For more details of the presentation, see Figure~\ref{fig:heatmapTime}.
}
\label{fig:heatmapLuby}
\end{figure*}

\newcommand{\pgfmanual}[0]{\tikzset{paint/.style={ draw=##1!50!black, fill=##1!50 },decorate with/.style={decorate,decoration={shape backgrounds,shape=##1,shape size=2.0mm}}}\tikz\draw[decorate with=diamond, paint=green!40!black] (-0.2,-0.1) -- (0.005,-0.1);}

In the following, we investigate different \emph{bet-and-run} strategies for different instances. 
For each combination of algorithm, instance, and overall run time budget, we average the outcomes of 100 independent repetitions on a compute cluster with Intel Xeon E5620 CPUs (2.4GHz).\footnote{Our code and results have been made publicly available: \url{https://bitbucket.org/markuswagner/restarts}} 
The benefit of using 100 repetitions is that the standard error of the mean is only 10\% of the sample's standard deviation, which means that the resulting averages are reasonably accurate representatives of the actual average performance, despite the algorithms' randomized nature. This in turn allows us to draw conclusions about the average advantage of our approach across different instances and problem domains. 

For each heatmap we use one fixed total time budget, and we systematically vary the number of runs in the first phase and their run times. In each plot we show a diagonal line that indicates the schemes \textsc{Restarts}$^{\text{x}}_{\text{1/x}}$, which corresponds to performing $x$ independent runs with each $1/x$-th of the total budget. Every scheme above this line would violate the total time budget, which gives the heatmaps a triangular shape.

Before we come to the discussion of the experimental results, a note regarding the implementation. Often algorithm implementations do not allow us to pause and continue their operation at arbitrary points in time, or to provide initial solutions together with a full internal state of the algorithm. While both options are implementable, the source code is not always available, or it cannot be easily modified. Therefore, we employ a trick that can easily be applied \emph{if} the implementation accepts run time limits and seeds for the random number generation. In our investigations, we first execute all runs of the first phase sequentially, each with the respective allotted time budget. Then we determine the best performing run $b$ that used time $t_b$ and the randomly set seed $s_b$. In order to complete the restart scheme's second phase, we run $b$ not just with the remaining time budget $t_2$ (see Figure~\ref{fig:illustration}), but we restart it from scratch with $t_b+t_2$ and with the previously used seed $s_b$. This allows the algorithm to reach its previous state after $t_b$ (which we do not count toward the total time budget) and to continue for $t_2$.

Figure~\ref{fig:heatmapTime} depicts how the total budget influences the relative performance of the restart strategies on a TSP instance. For a small total time budget we see that 4 to 	10 short initial runs are best; with an increase in the budget, more and more strategies with even longer initial run times perform better than the single-run strategy. Also, it should be noted that when the number of initial runs or the time budget for them increases too much, the performance of the scheme deteriorates quickly.

Similar observations also hold for our restart scheme on %on \MaxSAT and 
\MVC instances, as shown in Figure~\ref{fig:heatmapLuby}.
%Interestingly, in the case of the \MaxSAT instance, the schemes with a few but very long initial runs performs best, which is in contrast to what we observed on the TSP instance.
Also, we see that the Luby time unit has an impact on the overall performance of the approach, as does the length of the Luby sequence used. 

\definecolor{shadecolor}{rgb}{1,1,1}
\setlength{\fboxsep}{0pt}
\setlength{\fboxrule}{1pt}
\newcommand{\restartsLabel}[2]{\pgfmanual\tiny\rotatebox{26}{\hspace{-1mm}\colorbox{white}{\restarts{#1}{#2}}}}
\newcommand{\restartsLabelTwo}[2]{\pgfmanual\tiny\rotatebox{26}{\hspace{-1mm}\colorbox{white}{\restarts{#1}{#2}}}}
\begin{figure}[t]%
\centering%
\begin{overpic}[width=70mm,trim={0 0 0 0},clip,]%DONOTCHANGEwidth!!!requiresFixingTheRestartsLabelsWhenChanged
{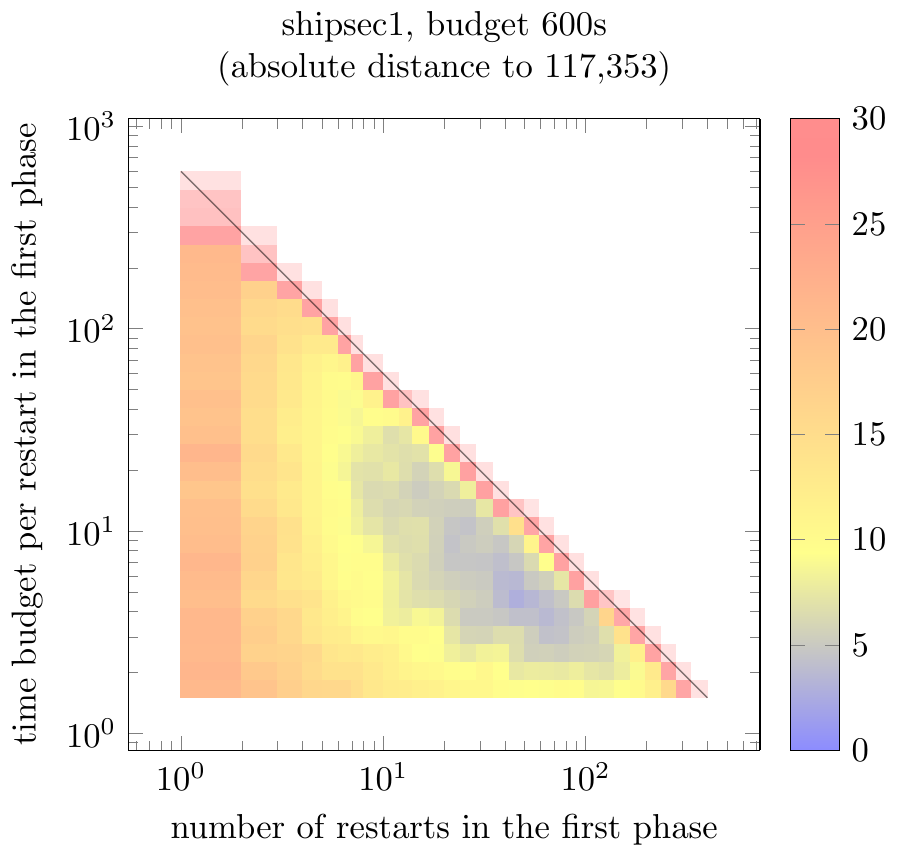}
%{{heatmap/151210mona-lisa100K/mona-lisa100K.tsp.600s.noluby.array-clusterout.csv}.pdf}
%\put(18,16){\pgfmanual \tiny\hspace{-1mm}\rotatebox{-15}{\restarts{1}{100}}}
\put(19,75){\restartsLabel{1}{100}}
\put(32,61.7){\restartsLabel{4}{25}}
\put(32,52){\restartsLabel{4}{10}}
\put(32,39){\restartsLabel{4}{2.5}}
\put(32,30){\restartsLabel{4}{1}}
\put(41,47){\tiny\rotatebox{26}{\colorbox{white}{\restarts{10}{4}}}}
\put(41,44){\pgfmanual}
\put(41,30){\restartsLabel{10}{1}}
\put(41,22){\restartsLabel{10}{0.4}}
\put(55,30){\restartsLabel{40}{1}}
\put(55,16){\restartsLabel{40}{0.25}}
\end{overpic}
\caption{Illustration of the representative restart strategies chosen in
Section~\ref{sec:ChoosingStrategies} and compared with each other
in Section~\ref{sec:final}.
Each strategy is marked with an \protect\pgfmanual.
The background shows the MVC instance shipsec1 ($400\cdot t_\text{init}$).
Note  that the entire leftmost column represents the scheme \restarts{1}{100}.
The strategy \restarts{$*$}{0.1} is not defined here 
due to the overall budget being $400\cdot t_\text{init}$.}
\label{fig:heatmapMona}
\end{figure}

In summary, no single \emph{bet-and-run} performs best across both problem domains. However, there are always schemes that outperform the naive scheme with just a single run, giving clear evidence for an advantage of our \emph{bet-and-run} approach. 

For our general study of restart strategies (across different problems and instances), we use the following diverse set of 14 strategies that vary in the number of runs used in Phase~1 and in the run time allocated to each run.

\begin{itemize}%[nosep]
\item Phase 1 takes 100\% of the total time
\begin{itemize}
\item \restarts{1}{100}: 1 regular run
\item \restarts{4}{25}: 4 runs, 25\% each
\end{itemize}
\item Phase 1 takes 40\% of the total time
\begin{itemize}
\item \restarts{4}{10}: 4 runs with 10\% each
\item \restarts{10}{4}: 10 runs with 4\% each
\item \restarts{40}{1}: 40 runs with 1\% each
\end{itemize}
\item Phase 1 takes 10\% of the total time
\begin{itemize}
\item \restarts{4}{2.5}: 4 runs with 2.5\% each
\item \restarts{10}{1}: 10 runs, each with 1\% each
\item \restarts{40}{0.25}: 40 runs, each with 0.25\% each
\end{itemize}
\item Phase 1 takes 4\% of the total time
\begin{itemize}
\item \restarts{4}{1}: 4 runs with 1\% each
\item \restarts{10}{0.4}: 10 runs with 0.4\% each
\item \restarts{40}{0.1}: 40 runs with 0.1\% each
\end{itemize}
\item Three Luby-based strategies
\begin{itemize}
\item \restartsluby{4}{1}: Luby sequence length 4\\(5 units in total)
\item \restartsluby{10}{1}: Luby sequence length 10\\(16 units in total)
\item \restartsluby{40}{1}: Luby sequence length 40\\(96 units in total)
\end{itemize}
\end{itemize}
In all cases, Phase 2 continues with the \emph{bet-and-run} found in Phase 1.

In Figure~\ref{fig:heatmapMona} we show for one representative instance the points in the \textsc{Restarts}-landscape that we will be investigating subsequently.

\pgfplotstableread{
tinit	c1	c2	c3	c4	c5	c6
100	147127.33	147120.61	147113.36	147117.49	147000.00	147124.13
400	147123.68	147113.89	147109.27	147111.01	147114.53	147111.09
1000	147123.39	147113.50	147108.89	147109.48	147110.59	147109.23
4000	147121.25	147112.91	147107.95	147106.04	147106.22	147106.58
10000	147120.31	147112.67	147104.57	147105.02	147105.60	147106.226
}\tinitstudy

\pgfplotscreateplotcyclelist{RESTARTScolors}{
    red,only marks,every mark/.append style={fill=red},thick,mark size=1.7,mark=*\\
    blue!80!black,only marks,every mark/.append style={fill=white},thick,mark size=1.5,mark=square*\\
    black,only marks,thick,mark size=2.3,mark=star\\
    green!60!black,only marks,every mark/.append style={fill=green!60!black},thick,mark=triangle*\\
    violet,only marks,every mark/.append style={fill=violet},thick,mark=diamond*\\
	orange,only marks,every mark/.append style={fill=white},thick,mark=pentagon*\\
    gray,only marks,every mark/.append style={fill=gray},mark size=2,mark=x\\%
}

\pgfplotsset{
cycle list name=RESTARTScolors,
    width=8cm,height=5cm,
    legend style={font=\small},
    legend pos=outer north east,
    tick label style={font=\small},
    label style={font=\small},
    title style={font=\small},
%axis line style={opacity=0}, % make axes disappear
tick pos=left, %only show bottom and left ticks
%xticklabels={},
%x tick style={white},
%y tick style={white},
ylabel style={align=center},
    xtick={100,400,1000,4000,10000},
    xticklabels={100,400,1000,4000,10000},
    xticklabel style={rotate=0, inner sep=1mm},
    scaled y ticks=false,
    ymin=147100,ymax=147128,
    yticklabel style={rotate=90, inner sep=1mm},
	yticklabels={147103,147115,147127},
	ytick={147103,147115,147127},
    y tick label style={
        /pgf/number format/.cd,
            fixed,
            fixed zerofill,
            precision=0,
        /tikz/.cd
    },
    enlargelimits=0.1,
}%

\begin{figure}[t]\centering
\begin{tikzpicture}
\begin{semilogxaxis}[ylabel={\hspace{4mm}minimum vertex cover size\\\hspace{4mm}(smaller = better)},xlabel={overall time budget $x*t^{init}$}]
\addplot+ table[x=tinit,y=c1]\tinitstudy;\label{plot:regularRun};
%\addlegendentry{regular run}
\addplot+ table[x=tinit,y=c2]\tinitstudy;\label{plot:25each};
%\addlegendentry{four runs 25\% each}
\addplot+ table[x=tinit,y=c3]\tinitstudy;\label{plot:10-4};
%\addlegendentry{\restarts{10}{4}}
\addplot+ table[x=tinit,y=c4]\tinitstudy;\label{plot:10-1};
%\addlegendentry{\restarts{10}{1}}
\addplot+ table[x=tinit,y=c5]\tinitstudy;\label{plot:10-0.4};
%\addlegendentry{\restarts{10}{0.4}}
\addplot+ table[x=tinit,y=c6]\tinitstudy;\label{plot:10-1 luby};
%\addlegendentry{\restartsluby{10}{1}}
\end{semilogxaxis}
\end{tikzpicture}
\caption{Solution quality achieved for sc-shipsec5 by different schemes for five different total time budgets. Shown are the averages of 100 independent runs of six different strategies: \ref*{plot:regularRun} one regular run, \ref*{plot:25each} four runs with 25\% of the time budget each, \ref*{plot:10-4} \restarts{10}{4}, \ref*{plot:10-1} \restarts{10}{1}, \ref*{plot:10-0.4} \restarts{10}{0.4}, and \ref*{plot:10-1 luby} \restartsluby{10}{4}. Note that \ref*{plot:10-0.4}  \restarts{10}{0.4} is not defined for $100\cdot t_\text{init}$.
It is also curious to observe that the best restart strategy \ref*{plot:10-4}
with time budget $100\cdot t_\text{init}$ outperforms a single-run strategy \ref*{plot:regularRun}
with time budget $10000\cdot t_\text{init}$.
}
\label{fig:tinitstudy}
\end{figure}

\begin{landscape}
\begin{figure*}[t]%
\centering%
\hspace{-2mm}%
\renewcommand{\arraystretch}{1.2}\small%
\begin{tabular}{l}
\bf Budget: $100\cdot t_\text{init}$\\
\restarts{1}{100}\\
\restarts{4}{25}\\
%\hline
\restarts{4}{10}\\
\restarts{10}{4}\\
\restarts{40}{1}\\
%\hline
\restarts{4}{2.5}\\
\restarts{10}{1}\\
%\restarts{40}{0.25}\\
%\hline
\restarts{4}{1}\\
%\restarts{10}{0.4}\\
%\restarts{40}{0.1}\\
%\hline
\restartsluby{4}{1}\\
\restartsluby{10}{1}\\
\restartsluby{40}{1}\\
\end{tabular}
\pgfplotstabletypeset[
    color cells={min=1.7,max=10.4},
    col sep=comma,
    /pgf/number format/fixed,fixed zerofill,precision=1,
    /pgfplots/colormap={whiteblue}{
[1cm]%
rgb255(0cm)=(140,140,255)
rgb255(1cm)=(255,255,140)
rgb255(2cm)=(255,198,140)
rgb255(3cm)=(255,140,140)
    }
]{
TSP
8.1
8.2
1.7
3
5.3
4.7
5
5.6
7.1
6.9
10.4
}%
\pgfplotstabletypeset[
    color cells={min=2.3,max=6.8},
    col sep=comma,
    /pgf/number format/fixed,fixed zerofill,precision=1,
    /pgfplots/colormap={whiteblue}{
[1cm]%
rgb255(0cm)=(140,140,255)
rgb255(1cm)=(255,255,140)
rgb255(2cm)=(255,198,140)
rgb255(3cm)=(255,140,140)
    }
]{
MVC
6.8
4
3.2
2.6
2.3
2.6
4.4
5.9
6.6
4.1
4.1
}%
\hspace{5mm}%
\renewcommand{\arraystretch}{1.2}\small%
\begin{tabular}{l}
\bf Budget: $400\cdot t_\text{init}$\\
\restarts{1}{100}\\
\restarts{4}{25}\\
%\hline
\restarts{4}{10}\\
\restarts{10}{4}\\
\restarts{40}{1}\\
%\hline
\restarts{4}{2.5}\\
\restarts{10}{1}\\
\restarts{40}{0.25}\\
%\hline
\restarts{4}{1}\\
\restarts{10}{0.4}\\
%\restarts{40}{0.1}\\
%\hline
\restartsluby{4}{1}\\
\restartsluby{10}{1}\\
\restartsluby{40}{1}\\
\end{tabular}
\pgfplotstabletypeset[
    color cells={min=2.3,max=12.3},
    col sep=comma,
    /pgf/number format/fixed,fixed zerofill,precision=1,
    /pgfplots/colormap={whiteblue}{
[1cm]%
rgb255(0cm)=(140,140,255)
rgb255(1cm)=(255,255,140)
rgb255(2cm)=(255,198,140)
rgb255(3cm)=(255,140,140)
    }
]{
TSP
12.3
3
3.7
2.3
3
6.2
5.6
7.7
9.9
9
10.8
10.3
7.2
}%
\pgfplotstabletypeset[
    color cells={min=1.3,max=9.4},
    col sep=comma,
    /pgf/number format/fixed,fixed zerofill,precision=1,
    /pgfplots/colormap={whiteblue}{
[1cm]%
rgb255(0cm)=(140,140,255)
rgb255(1cm)=(255,255,140)
rgb255(2cm)=(255,198,140)
rgb255(3cm)=(255,140,140)
    }
]{
MVC
9.4
5.4
4.4
1.8
1.3
5.2
3.2
3.7
6.5
4.5
6.4
3.5
2.5
}%
\hspace{5mm}
\renewcommand{\arraystretch}{1.2}\small%
\begin{tabular}{l}
\bf Budget: $1000\cdot t_\text{init}$\\
\restarts{1}{100}\\
\restarts{4}{25}\\
%\hline
\restarts{4}{10}\\
\restarts{10}{4}\\
\restarts{40}{1}\\
%\hline
\restarts{4}{2.5}\\
\restarts{10}{1}\\
\restarts{40}{0.25}\\
%\hline
\restarts{4}{1}\\
\restarts{10}{0.4}\\
\restarts{40}{0.1}\\
%\hline
\restartsluby{4}{1}\\
\restartsluby{10}{1}\\
\restartsluby{40}{1}\\
\end{tabular}
\pgfplotstabletypeset[
    color cells={min=1.9,max=13.7},
    col sep=comma,
    /pgf/number format/fixed,fixed zerofill,precision=1,
    /pgfplots/colormap={whiteblue}{
[1cm]%
rgb255(0cm)=(140,140,255)
rgb255(1cm)=(255,255,140)
rgb255(2cm)=(255,198,140)
rgb255(3cm)=(255,140,140)
    }
]{
TSP
13.7
5.1
5.6
1.9
2.2
6.9
6.0
7.1
10.0
9.9
11.0
12.8
9.8
3.0
}%
\pgfplotstabletypeset[
    color cells={min=1,max=10.1},
    col sep=comma,
    /pgf/number format/fixed,fixed zerofill,precision=1,
    /pgfplots/colormap={whiteblue}{
[1cm]%
rgb255(0cm)=(140,140,255)
rgb255(1cm)=(255,255,140)
rgb255(2cm)=(255,198,140)
rgb255(3cm)=(255,140,140)
    }
]{
MVC
10.1
6.9
6.6
2.8
1.0
5.8
2.4
2.5
5.9
3.8
4.9
7.0
3.4
2.2
}%
\tikz[overlay,remember picture]% 1000
\foreach \x in  {1,4,7,10} %{0,1,...,12}%
\draw[white,line width=2pt] (-2.9,2.6-\x*0.457) -- ++(3.,0);
\tikz[overlay,remember picture]%
\foreach \x in {1,4,7,9}% {0,1,...,12}% 500
\draw[white,line width=2pt] (-9.5,2.35-\x*0.457) -- ++(3.,0);
\tikz[overlay,remember picture]%
\foreach \x in {1,4,6,7}% {0,1,...,12}% 500
\draw[white,line width=2pt] (-16.3,1.9-\x*0.457) -- ++(3.,0);
\caption{Average rank (smaller values are better)
of the different restart strategies for the two optimization problems with
three total time budgets. 
Strategies that use the same total time for the first phase are grouped together, as are the ones based on the Luby sequence. 
The colors correspond to the average rank of a scheme (colder colors are better).
The two \emph{bet-and-run} strategies \restarts{10}{4} and  \restarts{40}{1} have the best average rank. 
A single-run with no restarts has the worst average rank.
}
\label{fig:crossStudy}
\end{figure*}
\end{landscape}

\section{Cross Problem Study}
\label{sec:final}

A crucial decision is the \emph{total time budget} allotted for each instance.
If the time limit is too short, no strategy has enough time to finish even its
initialization.  If the time limit is too long, the differences between the strategies
might vanish. To investigate the impact of different total run time budgets, we consider five different budgets. These budgets are all relative to the time $t_\text{init}$ needed to initialize the algorithm with the given instance.
The overall run time budgets that we consider are $100\cdot t_\text{init}$, $400\cdot t_\text{init}$, $1\,000\cdot t_\text{init}$, $4\,000\cdot t_\text{init}$, and $10\,000\cdot t_\text{init}$.

As an example we present Figure~\ref{fig:tinitstudy} to show the results of the impact of the total run time budget, considering the minimum vertex cover instance sc-shipsec5. It is clearly visible that a single run without restarts has the worst performance. This configuration is outperformed by all others, even the one where four independent runs are given just 25\% of the total computation budget. These observations hold independently of the chosen total budget. 
When relatively little time is available (e.g. $100\cdot t_\text{init}$), the performance of the different restart schemes varies significantly.However, the differences between our different schemes seem to disappear with increasing time budget, and the restart schemes are able to find smaller and smaller vertex covers.

In order to allow all restart strategies introduced in Section~\ref{sec:ChoosingStrategies}
to have a fair chance to finish at least the initialization in each run, we have to make sure
that each run of Phase~1 gets at least time $t_\text{init}$. For example for
\restarts{40}{0.1} this implies that the total time budget has to be at least $1000\cdot t_\text{init}$.
As computational resources are the bottleneck for our subsequent studies,
we focus on the shortest three total time budgets: $100\cdot t_\text{init}$, $400\cdot t_\text{init}$ and $1000\cdot t_\text{init}$. % (for time budget $400\cdot t_\text{init}$ we omit \restarts{40}{0.1}).

In our first \emph{cross problem domain study} 
we determine the average rank of the 14~restart strategies described in Section~\ref{sec:ChoosingStrategies} for the two optimization problems. 
For each of the 20 instances 
listed in Section~\ref{sec:ProblemsAndBenchmarks} we perform 100 independent repetitions.
Based on the results, we then determine the relative ranks of the 14~restart strategies for both total time budgets.

Figure~\ref{fig:crossStudy} shows the average ranks% for all instances of each problem
, which are reflecting the trends that we have previously seen in the heatmaps. 
For the two different problem domains, we observe the following:
\begin{itemize}%[nosep]
\item For TSP it is best to use a relatively large fraction (40\%) of the total time budget with 4 to 40 runs in the first phase. If less time is used for the initial runs, then the average rank worsens quickly.
\item For the MVC instances, the range of effective budgets for the first phase is wider, it covers the range from 4--40\%. However, schemes with only a few runs perform the worst. 
\end{itemize}

In all cases, 
the \emph{bet-and-run} approaches clearly outperform the commonly used single-run strategy, which ends up on one of the worst ranks. 
When considering the average performance across all total time limits, our schemes \restarts{40}{1} and \restarts{10}{4} can almost be considered universal for the given instances and solvers. For the TSP and the MVC, they achieve the best or second best rankings. 
The universal sequence of \citet{luby1993}
turned out inferior compared to restarts of fixed length, which matches the earlier studies on the decision version of SAT/UNSAT problems by~\citet{Audemard2012restartsSAT}.

Lastly, we investigate the broader applicability of the best performing strategy \restarts{40}{1} when the total time limit was $1000\cdot t_\text{init}$. We apply it to the 86 MVC instances used in \cite{cai2015fastvc}, which come from 10 categories of networks, and to the 111 symmetric TSP instances from TSPlib, which cover geographical instances as well as circuit board layouts. As before, we repeat each experiment 100 times independently in order to get reasonable estimates of the performance distribution. 
The results in Figure~\ref{fig:pies} show that on almost all instances, the standard run is outperformed by our bet-and-run strategy \restarts{40}{1}. 

\newcommand{\fancypie}[5]{\def\angle{0}
\def\radius{0.9}
\def\cyclelist{{"red","green","black!15!white","black!40!white"}}
\newcount\cyclecount \cyclecount=-1
\newcount\ind \ind=-1
\begin{tikzpicture}[nodes = {font=\sffamily}]
  \foreach \percent/\name in {
      #2/worse,
      #3/better,
      #4/identical,
      #5/insignificant
    } {
      \ifx\percent\empty\else               % If \percent is empty, do nothing
        \global\advance\cyclecount by 1     % Advance cyclecount
        \global\advance\ind by 1            % Advance list index
        \ifnum3<\cyclecount                 % If cyclecount is larger than list
          \global\cyclecount=0              %   reset cyclecount and
          \global\ind=0                     %   reset list index
        \fi
        \pgfmathparse{\cyclelist[\the\ind]} % Get color from cycle list
        \edef\color{\pgfmathresult}         %   and store as \color
        % Draw angle and set labels
        \draw[fill={\color!50},draw={\color}] (0,0) -- (\angle:\radius)
          arc (\angle:\angle+\percent*3.6/#1*100:\radius) -- cycle;
        \node at (\angle+0.5*\percent*3.6/#1*100:0.7*\radius) {\percent};
        %\node[pin=\angle+0.5*\percent*3.6*100/#1:\name]
         % at (\angle+0.5*\percent*3.6*100/#1:\radius) {};
        \pgfmathparse{\angle+\percent*3.6/#1*100}  % Advance angle
        \xdef\angle{\pgfmathresult}         %   and store in \angle
      \fi
    };
\end{tikzpicture}
}

\begin{figure}
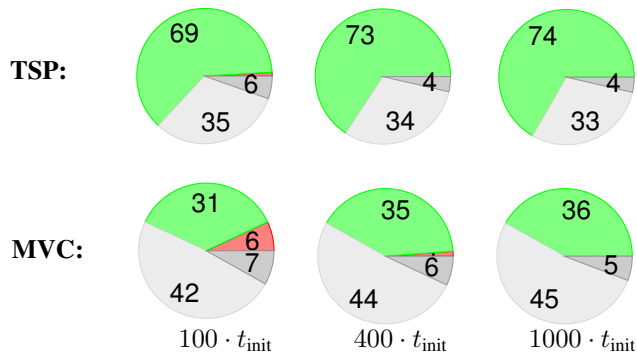
\centering%
\noindent
\begin{minipage}{1.3cm}\textbf{TSP:}\\\vspace*{15mm}\end{minipage}
\fancypie{111}{1}{69}{35}{6}
\hspace*{.3mm}
\fancypie{111}{0}{73}{34}{4}
\hspace*{1mm}
\fancypie{111}{0}{74}{33}{4}
\vspace*{-7mm}
\\
%\noindent
\begin{minipage}{1.2cm}\textbf{MVC:}\\\vspace*{15mm}\end{minipage}
\fancypie{86}{6}{31}{42}{7}
\fancypie{86}{1}{35}{44}{6}
\fancypie{86}{0}{36}{45}{5}
\vspace*{-10mm}
\\
\hspace*{20mm}
$100\cdot t_\text{init}$
\hspace*{9mm}
$400\cdot t_\text{init}$
\hspace*{9mm}
$1000\cdot t_\text{init}$
\caption{Statistical comparison of \restarts{40}{1} and \restarts{1}{100} (no restarts) with Wilcoxon rank-sum test (significance level $p=0.05$) for both problems and three total time budgets.
The colors have the following meaning:
Green indicates that \restarts{40}{1} is statistically better,
Red indicates that \restarts{40}{1} is statistically worse,
Light gray indicates that both performed identical,
Dark gray indicates that the differences were statistically insignificant.
Overall, the solutions for most problem instances were either improved or stayed
unchanged by introducing our bet-and-run strategy. 
Worsenings due to our bet-and-run strategy do not occur for the
largest time budget ($1000\cdot t_\text{init}$).
Within the medium time budget ($400\cdot t_\text{init}$),
0~out of 111~TSP-instances and 1~out of 86~MVC-instances got worse.
Within the smallest time budget ($100\cdot t_\text{init}$),
1~out of 111~TSP-instances and 6~out of 86~MVC-instances got worse.
}
\label{fig:pies}
\end{figure}

%\newpage
\section{Conclusions and Future Work}

We study a generic \emph{bet-and-run} restart strategy, which is easy to implement as an additional speed-up heuristic for solving difficult optimization problems. We demonstrate its efficiency on two
classical NP-complete optimization problems with \emph{state-of-the-art solvers}. Our experiments show a\ignore{ statistically} significant advantage of \emph{bet-and-run} strategies on all problems.
The best strategy overall was \restarts{40}{1}, which in the first phase does
40~short runs with a time limit that is 1\% of the
total time budget and then uses the remaining 60\% of
the total time budget to continue the best run of the first phase.
The universal sequence of \citet{luby1993} turned out inferior.

The gain achieved by our \emph{bet-and-run} strategy differs depending on the studied optimization problem.
For both TSP and MVC the gain is significant.

As the two problem domains are structurally different,
we expect that \emph{bet-and-run} strategies are generally helpful.
Future research should study further classes of optimization problems
such as multi-objective problems or continuous domains.
While we focus on strategies with two phases only, it is interesting
to consider iterated or hierarchical best-of strategies.
Another direction are dynamic \emph{bet-and-run} strategies, which
restart runs that stop improving.

\newpage 

\bibliographystyle{myabbrvnat}
\bibliography{restarts}

\end{document}